\documentclass[12pt]{article}

\usepackage[margin=1in]{geometry}
\usepackage{amsmath}
\usepackage{graphicx}
\usepackage{hyperref}
\usepackage{url}
\usepackage{caption}
\usepackage{subcaption}
\usepackage{float}
\usepackage[numbers,sort&compress]{natbib}

\title{\textbf{TheBlueScrubs-v1, a comprehensive curated medical dataset derived from the internet}}

\author{
    Luis Felipe$^{1,3}$,
    Carlos Garcia$^{1,3}$,
    Issam El Naqa$^{1,3}$,
    Monique Shotande$^{1,3}$,\\
    Aakash Tripathi$^{1,3}$,
    Vivek Rudrapatna$^{2,3}$,
    Ghulam Rasool$^{1,3}$,\\
    Danielle Bitterman$^{4}$,
    Gilmer Valdes$^{1,3}$\\[6pt]
    $^{1}$Machine Learning Department, Moffitt Cancer Center, Tampa, Florida.\\
    $^{2}$Center for Real World Evidence, UCSF, San Francisco, California.\\
    $^{3}$The Blue Scrubs, Tampa, Florida.\\
    $^{4}$Harvard Medical School, Boston, Massachusetts.
}
\date{\vspace{-5ex}}

\begin{document}

\maketitle

\noindent \textbf{Corresponding author:} \texttt{luis.felipe@moffitt.org}

\begin{abstract}
The need for robust and diverse data sets to train clinical large language models (cLLMs) is critical given that currently available public repositories often prove too limited in size or scope for comprehensive medical use. While resources like PubMed provide foundational medical literature, they capture only a narrow range of formal publications and omit the broader medical discourse on the internet. To address these deficits, we introduce TheBlueScrubs-v1, a curated dataset of over 25 billion medical tokens—nearly three times larger than PubMed—drawn from a broad-scale internet corpus. Our two-stage filtering pipeline employs a fast Logistic Regression model for document screening (achieving an AUC of approximately 0.95 on external validation), followed by verification via a 70B-parameter Llama 3.1 instruct model. Each text is assigned three LLM-based quality scores encompassing medical relevance, precision/factual detail, and safety/ethical standards. Clinician reviews confirm high concordance with these automated evaluations, and a specialized cancer classifier further labels approximately 11 billion oncology tokens. Two demonstration tasks highlight the dataset’s practical value: first, we distill the safety evaluations to a smaller BERT-style model that reaches an AUC near 0.96 on unseen data; second, we fine-tune a compact LLM on a filtered subset, showing measurable improvements over standard baselines in medical benchmarks. This Data Descriptor details the dataset’s creation and validation, underscoring its potential utility for medical AI research.
\end{abstract}

\section{Background \& Summary}

The development of clinical large language models (cLLMs)—language models designed for tasks involving clinical language and patient-oriented discourse—requires access to large and diverse training corpora. However, public repositories of medical text remain relatively small. General-purpose datasets like Common Crawl\textsuperscript{1} have been instrumental in scaling up LLMs, and recent work demonstrates their utility for domain-specific applications as well\textsuperscript{2}. For instance, a study by DeepSeek\textsuperscript{2} curated and filtered mathematical content within Common Crawl to create state-of-the-art mathematical and coding reasoning models—underscoring the value of large, heterogeneous datasets for specialized tasks.

While PubMed\textsuperscript{3} is an important and freely available source of trustworthy biomedical text, its scope is limited to (1) peer-reviewed journal publications, (2) domain-specific scientific content often focused on molecular, biochemical, or genomic research, and (3) a total size of approximately 10 billion tokens\textsuperscript{3}. This scale is insufficient for training very large language models from scratch, which, according to current scaling laws, typically require hundreds of billions to trillions of tokens\textsuperscript{4}. Incorporating other publicly available medical datasets (e.g., clinical guidelines\textsuperscript{4}) does not fully overcome this limitation. For example, the Meditron suite\textsuperscript{4} was trained with a total of 70 billion tokens, and DeepSeekMath 7B\textsuperscript{5} used 120 billion tokens—both still far from the trillion-token scale needed for comprehensive pretraining.

To address this gap, we introduce \textbf{The Blue Scrubs} (\url{https://huggingface.co/datasets/TheBlueScrubs/TheBlueScrubs-v1})—a curated dataset specifically for medical LLM development. It is extracted from SlimPajama\textsuperscript{6}, a 627-billion-token deduplicated version of RedPajama\textsuperscript{7}, comprising Common Crawl, C4, GitHub, books, arXiv, Wikipedia, and StackExchange. From SlimPajama, we isolate approximately 25 billion tokens of clearly medical content by applying a trained classifier with a probability threshold of $\geq0.8$ to balance filtering quality and corpus size. This is more than twice the size of PubMed, providing not only depth in formal medical literature but also breadth across informal, real-world medical discussions—such as those found in patient forums and health-related web content.

We release this dataset as TheBlueScrubs-v1, which includes all documents scoring $\geq0.8$ on our medical probability metric. This metric was generated using a linear classifier with bag-of-words features, as described in detail in the Methods section. Each document was then scored by a strong open-source large language model, Llama 3.1 (70B)\textsuperscript{9}, along three evaluation axes: scope of medical relevance, precision and factual detail, and safety and ethical standards. These LLM-based scores were cross-validated by a team of clinicians and GPT-4o-2024-08-06\textsuperscript{8}, demonstrating high concordance. For each text, we provide its medical probability score (ranging from 0.8 to 1.0), the full SlimPajama source indicating which original subset the text was drawn from, three LLM-based quality scores (on a 1--5 scale), and a cancer vs. non-cancer label indicating oncology-related content.

Altogether, TheBlueScrubs-v1 comprises approximately 25 billion tokens from 11.5 million documents, each richly annotated. Table 1 summarizes the distribution of texts by their original RedPajama source after applying our data pipeline.

\subsubsection*{Potential Use Cases}

TheBlueScrubs-v1 enables several downstream applications. One central use case is domain-specific model building and synthetic data generation. Researchers can fine-tune large language models on the high-scoring subset of medical texts to develop clinically specialized LLMs. In addition, the dataset can be used in synthetic data pipelines—rewriting or augmenting raw medical content—to reduce noise, improve coherence, and enhance efficiency in training pipelines for real-world medical applications.

Another application is misinformation detection and safety testing. The inclusion of structured metadata such as LLM-based scores on medical relevance, factual detail, and safety enables the creation of automated systems for identifying misleading or harmful content. Furthermore, the dataset supports red-teaming exercises to stress-test models against adversarial prompts involving controversial or ethically sensitive topics. These tests can be used to systematically probe model vulnerabilities and improve safeguards in clinical deployment environments.

\section{Methods}

This section describes the technical steps involved in the construction of The Blue Scrubs dataset, including the initial filtering using a logistic regression classifier, followed by large-scale LLM-based quality evaluation and a specialized cancer classification pipeline.

\subsection{Filtering Process (Linear Classifier)}

To distinguish medical from non-medical texts within SlimPajama, we trained a logistic regression classifier using TF-IDF features. The training set consisted of a balanced dataset of 60,000 samples—30,000 labeled as medical and 30,000 as non-medical. The labeling was performed by sampling documents from trusted medical sources and sources known to be general-purpose or unrelated to healthcare. Details of the composition and source attribution for these samples are described in Tables 2 and 3.

The choice of a linear model was motivated by its practicality and performance at large scale. It offered significantly faster inference times than transformer-based classifiers and demonstrated comparable discrimination ability in our internal benchmarks. In particular, our logistic regression classifier was capable of processing the entire 627 billion tokens of SlimPajama in under 24 hours using 8 NVIDIA H100 GPU cores. By contrast, preliminary experiments with a BERT-based classifier—specifically Med-BERT\textsuperscript{12}—suggested a projected runtime of approximately nine days for the same task, with no measurable performance gain. While these internal benchmarks are unpublished, we observed an area under the curve (AUC) of approximately 0.95 for both models when evaluated on the same test set, justifying the use of the logistic regression model due to its efficiency.

For evaluation, we conducted two primary tests. First, we used an 80-20 split of our internal dataset to construct a held-out test set. Second, we evaluated the classifier on a publicly available external benchmark from Hugging Face (\url{https://huggingface.co/datasets/ai-maker-space/medical_nonmedical}). This external dataset is composed of documents explicitly labeled as medical or non-medical by the dataset creators based on content and source metadata, although we note that its labeling methodology has not been independently audited. In both evaluations, the classifier demonstrated strong performance with AUC scores exceeding 0.95, as shown in Table 4.

Although the logistic regression classifier can operate on entire documents, we opted to split each input into 512-token fragments to enable a more granular evaluation and to ensure alignment with the evaluation setup used for later comparisons with transformer-based models. For documents consisting of multiple fragments, we aggregated the individual fragment scores to compute a final document-level probability score. While this process may not be required for linear models, it ensures consistency in the pipeline across models and downstream evaluation steps.

\subsubsection*{Threshold-Based Filtering for Further Processing}

After scoring all documents in SlimPajama, we manually inspected the quality of texts across the range of predicted medical probabilities. While a score greater than 0.5 generally indicated medical content, many documents in the 0.5--0.8 range included low-quality, tangential, or ambiguous material. Based on these observations, we set a more conservative threshold of $\geq0.8$ for downstream processing to ensure the retained corpus contained consistently high-quality and clearly medical content. Applying this threshold resulted in approximately 25.1 billion tokens, representing roughly 4\% of SlimPajama, spanning about 11.5 million texts. These documents were then subjected to a more detailed quality assessment using a large open-source LLM.

\subsection{Quality and Safety Evaluation with Llama 3.1 (70B)}

To further evaluate the quality of the filtered corpus, we used the Llama 3.1 (70B) Instruct model\textsuperscript{9} to score each document along three dimensions: (1) scope of medical relevance, (2) precision and factual detail, and (3) safety and ethical standards. Each category was scored on a 1--5 scale, with 5 indicating the highest quality or adherence. The evaluation was performed using vLLM\textsuperscript{13} across 8 H100 GPUs, enabling efficient large-batch inference. The inference was conducted with a temperature of 0.7, top-p sampling of 0.95, and a maximum of 128 output tokens per prompt. Each input could contain up to 4096 tokens, and the batch size was set to 5120. Log probability outputs were enabled to facilitate scoring.

Initially, we attempted to extract the 1--5 ratings from model output using regular expression parsing. However, inconsistencies in the generated text led to a $\sim20\%$ failure rate in score extraction. To improve reliability, we adopted a log probability--based scoring method. We forced numeric token outputs, recorded the log probabilities of tokens corresponding to the five score options, and converted these into normalized probabilities. The final score for each dimension was computed as the expected value across the 1--5 range and truncated to two decimal places. This method ensured consistency in scoring and reduced variance introduced by generation format variability.

The LLM-based scoring framework was validated using a random subset of 60 texts. These documents were independently reviewed by three clinicians using a simplified schema of Low, Neutral, and High for two dimensions—scope of medical relevance and safety and ethical standards. Precision and factual detail were excluded from human review due to the difficulty of consistent evaluation without reference material. The same subset was also scored using GPT-4o (OpenAI)\textsuperscript{8}, using identical prompts to those used for Llama 3.1. The results from Llama 3.1 demonstrated strong concordance with both clinician and GPT-4o assessments. Minor discrepancies occurred primarily in highly technical or borderline cases. The quantitative results from this comparison are presented in Figures 4--7 and discussed in the Technical Validation section.

\subsection{Cancer Classification}

While TheBlueScrubs-v1 is designed as a general-purpose medical dataset, we observed that a substantial portion of the filtered content pertained to oncology. To systematically annotate these texts, we implemented an additional classification pipeline focused on cancer-related content.

We began by identifying documents that included at least one mention of a predefined set of oncology-related keywords, such as ``cancer,'' ``tumor,'' ``oncology,'' ``carcinoma,'' ``neoplasm,'' ``metastasis,'' ``malignant,'' ``chemotherapy,'' ``radiotherapy,'' ``sarcoma,'' ``lymphoma,'' ``leukemia,'' and ``melanoma.'' This keyword-based selection served as a proxy label for training a second logistic regression classifier. A total of 60,000 documents were sampled from the previously filtered medical corpus—30,000 with evidence of keyword matches and 30,000 without. These documents were used to train a cancer-specific classifier using the same TF-IDF feature representation.

Unlike the initial keyword filter, the trained classifier generalized beyond the original keyword list and captured a broader range of oncology-related language. This included mentions of cancer treatment protocols, rare neoplasms, and indirect references to oncology (e.g., genomic mutations, metastasis risk). This was confirmed via manual inspection of the top-ranked cancer texts, which contained rich and varied clinical language beyond the seed terms.

The trained cancer classifier was applied to the full 25.1 billion-token medical corpus. Documents identified as cancer-related were annotated accordingly. The resulting subset contained approximately 11 billion tokens labeled as oncology-related, making it one of the largest open corpora for cancer language modeling to date. Table 6 presents the distribution of cancer vs. non-cancer texts, while Tables 7 and 8 summarize the LLM-based safety and quality scores for the cancer-specific subset.

\section{Data Records}

The Blue Scrubs dataset is publicly available at \url{https://huggingface.co/datasets/TheBlueScrubs/TheBlueScrubs-v1}. The dataset includes approximately 11.5 million documents, each annotated with a set of metadata fields generated through our filtering and evaluation pipeline. For each document, we provide a medical probability score ranging from 0.8 to 1.0, derived from our linear classifier. In addition, each document includes three evaluation scores produced by the Llama 3.1 (70B) model: scope of medical relevance, precision and factual detail, and safety and ethical standards—each rated on a 1 to 5 scale.

Additional metadata include the original SlimPajama source from which the document was drawn, allowing for stratification and analysis by domain (e.g., Common Crawl, GitHub, or academic repositories). Furthermore, each document is labeled as either cancer-related or not, based on a secondary classifier trained specifically to identify oncology-related language. The final dataset consists of 25.1 billion tokens in total, of which approximately 11 billion tokens were identified as pertaining to cancer topics.

\begin{figure}[H]
\centering
\includegraphics[width=\textwidth]{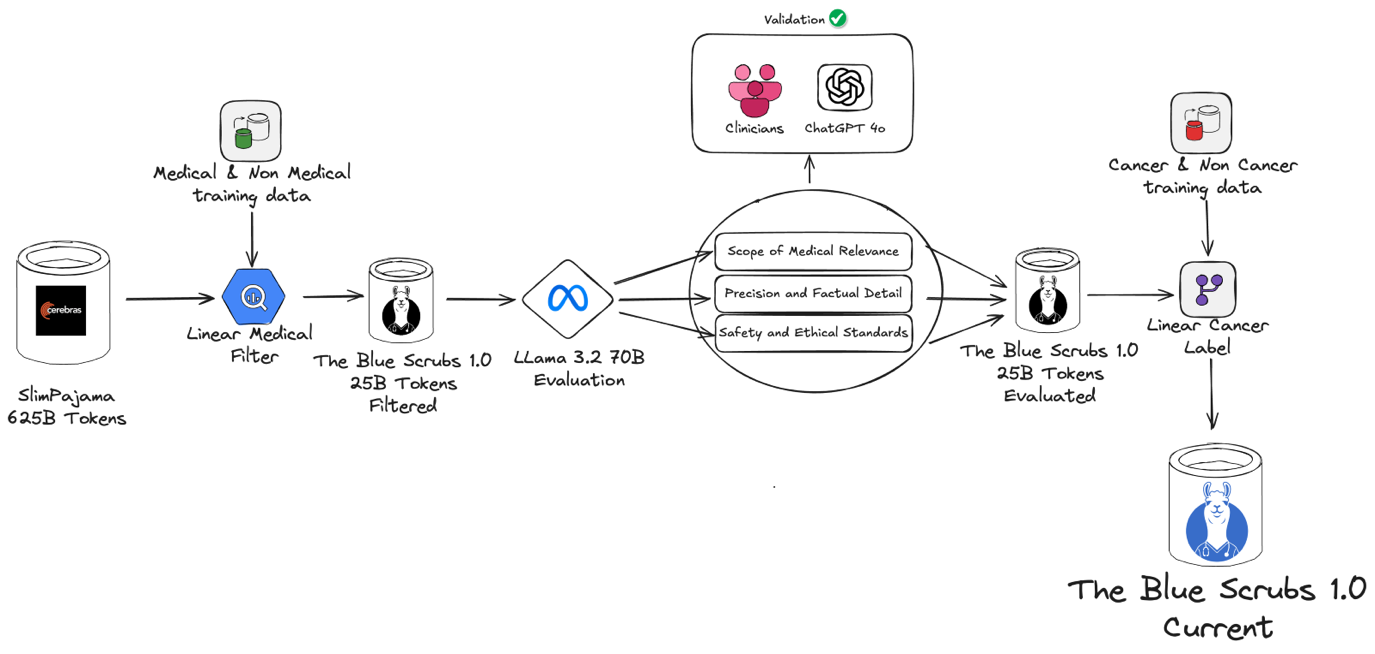}
\caption{The Blue Scrubs creation process through filtering, evaluation, and labeling.}
\label{fig:1}
\end{figure}

The dataset is organized in a sharded format, with each shard containing a segment of text and its associated metadata. A separate metadata file maps each document to its medical probability score, LLM evaluation scores, SlimPajama source, and cancer classification. Figure \ref{fig:1} provides a visual overview of the Blue Scrubs data pipeline, including the filtering and labeling stages, and Table \ref{tab:1} summarizes the distribution of texts across the various RedPajama source domains.

\section{Technical Validation}

The linear classifier used to separate medical from non-medical documents served as the foundational filter for TheBlueScrubs-v1. As described earlier, this model was trained on a balanced dataset of 60,000 examples and achieved high performance on both internal and external benchmarks. Table \ref{tab:4} presents the performance metrics for this classifier, confirming its ability to robustly distinguish medical content with an AUC exceeding 0.95. Only texts that scored above the threshold of 0.8 on this classifier—approximately 4\% of the total SlimPajama corpus—were retained for subsequent evaluation. This subset, which we refer to as the retained corpus, comprises roughly 25.1 billion tokens across 11.5 million documents and forms the basis of The Blue Scrubs dataset.

To validate the reliability of the large language model scoring framework, we compared the outputs from Llama 3.1 (70B) to human expert annotations and GPT-4o assessments. A randomly sampled set of 60 documents was independently rated by three clinicians along two dimensions: scope of medical relevance and safety and ethical standards. Each clinician used a simplified rating scale (Low, Neutral, High), while GPT-4o and Llama 3.1 generated scores on the original 1--5 scale. Figures \ref{fig:4}--\ref{fig:7} show the distributions of scores across all raters.

For medical relevance, the Llama 3.1 scores showed a largely monotonic relationship with both clinician and GPT-4o evaluations, suggesting that the model’s numerical scores align well with expert judgment in this dimension. However, the agreement was less clearly monotonic in the safety and ethical standards category. To further quantify model reliability, we examined inter-rater agreement between GPT-4o and Llama 3.1 across the 1--5 scale, with particular attention to disagreement at the extremes of the distribution. While we observed general alignment between model predictions and external raters, a small number of safety-critical cases—defined as those receiving a score of $\leq2$ by at least one rater—were missed by Llama 3.1. These edge cases highlight the importance of incorporating additional safety reviews or ensemble scoring when using LLMs to evaluate clinical content. Future work may benefit from explicitly quantifying false negatives in the safety domain and assessing the risk they pose in real-world applications.

Figure \ref{fig:3} shows the distribution of Llama 3.1 evaluation scores across the entire retained corpus. Approximately 78\% of the texts received a score of 5.0 for medical relevance, 68\% scored 5.0 for precision and factual detail, and 62\% scored 5.0 for safety and ethical standards. These results suggest that the majority of the retained dataset contains high-quality and clinically appropriate content, with good representation of medically detailed and ethically sound language.

In addition to general validation, we also assessed the quality and accuracy of the cancer classification pipeline. Table \ref{tab:6} presents the distribution of cancer versus non-cancer labels within the final dataset. Additional details about the average safety and quality scores for the cancer-labeled subset are shown in Tables \ref{tab:7} and \ref{tab:8}. The consistently high scores in this subset suggest that oncology-related documents—representing nearly one-third of the total dataset—are both abundant and of high quality. This reinforces the dataset’s value for downstream modeling tasks focused specifically on cancer, such as treatment recommendation or clinical summarization in oncology.

\section{Usage Notes}

To demonstrate the practical utility of TheBlueScrubs-v1, we present two use cases that show how the dataset can be applied to train clinically relevant models. These examples illustrate how high-quality, curated medical data can enhance model performance in both safety classification and general clinical language modeling.

\subsubsection*{First Use Case: Safety Classification}

In the first use case, we trained a lightweight safety classifier using a model based on the ModernBERT architecture. ModernBERT is a BERT-style transformer with 149 million parameters that incorporates Rotary Positional Embeddings, local-global alternating attention, and Flash Attention. Notably, it supports extended context lengths up to 8,192 tokens, which is particularly useful in clinical contexts involving longer documents. For consistency, we constrained the training to a 4,096-token context, matching the input size used in Llama 3.1 evaluations.

A balanced training dataset was constructed by sampling documents from TheBlueScrubs-v1 with safety scores at or below 2 and pairing them with an equal number of documents scoring above 2, based on Llama 3.1's evaluations. This resulted in a dataset of 83,636 labeled examples. Using knowledge distillation, we trained ModernBERT to predict the continuous Llama 3.1 safety score. On an out-of-sample test set released via Hugging Face, ModernBERT achieved a mean squared error (MSE) of 0.489 and a root mean squared error (RMSE) of 0.699 for continuous score prediction. When the task was recast as binary classification—labeling texts as unsafe (score $\leq2$) or safe (score $>2$)—the model attained an AUC of 0.9642. ROC curves at various thresholds are presented in Figure \ref{fig:8}. The trained classifier is publicly available on Hugging Face and can be readily adopted or adapted by researchers working on medical safety evaluation tasks.

\subsubsection*{Second Use Case: Clinical Language Model Fine-Tuning}

In the second use case, we fine-tuned a smaller Llama 3.1 model (8B parameters) using a high-quality subset of TheBlueScrubs-v1. From the full corpus, we selected a 2-billion-token subset by removing documents with safety scores below 4 and prioritizing those with high medical relevance. This filtered data emphasized content that is both clinically pertinent and ethically robust. For baseline comparison, we trained a second Llama 3.1 (8B) model on 2 billion tokens from UMLS—a well-established medical dataset.

To evaluate the fine-tuned models, we used multiple-choice questions drawn from the MMLU medical benchmark, as well as additional questions generated by GPT to assess domain-specific knowledge. The model trained on The Blue Scrubs subset matched or outperformed the UMLS-trained model across a range of tasks. Full results are presented in Table \ref{tab:9}. These findings suggest that TheBlueScrubs-v1 can serve as a competitive resource for training high-performing clinical LLMs, even when used in relatively small, curated quantities.

\section{Discussion}

TheBlueScrubs-v1 aims to bridge the gap in available large-scale medical datasets. By combining logistic regression–based classification and LLM-based evaluation, we filter and enrich medical content at scale. Our pipeline balances computational efficiency and domain specificity, producing a corpus where each text is annotated with medical relevance scores, factual precision scores, safety scores, and an oncology-specific label.

The dataset also exhibits key properties that make it valuable for medical AI development. First, it retains high clinical relevance, as established through both classifier-based and LLM-based filtering. Second, approximately one-third of the dataset focuses on oncology, providing a robust foundation for cancer-specific modeling. Third, the inclusion of Llama 3.1–based safety scores offers a transparent and interpretable measure of ethical quality across the corpus.

Despite these strengths, some limitations remain. As with any web-derived corpus, residual bias, misinformation, or outdated content may persist, even after filtering. Additionally, the LLM-based scoring process is computationally intensive; running the full pipeline on 25 billion tokens required 27 days using 8 H100 GPUs. Finally, while the original data sources comply with GDPR and local data protection regulations (notably through Common Crawl's policies), users are advised to exercise caution, especially when deploying models in clinical settings.

Looking ahead, we plan to extend The Blue Scrubs pipeline to the FineWeb corpus, a 15-trillion-token English-language web dataset that has been cleaned and deduplicated from Common Crawl. As with TheBlueScrubs-v1, we will apply domain-specific filtering strategies to isolate medical content, and we will continue using our ModernBERT-based evaluators to efficiently rate documents for safety and relevance. In parallel, we aim to develop finer-grained classifiers for clinical subdomains, including cardiology, neurology, and oncology, with the goal of producing even more targeted and modular datasets.

If significant non-English medical content is identified within FineWeb, we will also pursue multilingual expansion, adapting our filtering and evaluation procedures accordingly. Future comparisons between The Blue Scrubs framework and other medical corpora—such as PubMed and UMLS—will help refine our approach, with a focus on iterative improvement guided by user feedback and empirical validation. Through these efforts, we aim to contribute to the ongoing development of high-quality, scalable, and ethically sound datasets for clinical large language models.

\section{Code Availability}

All classifier code—including logistic regression and TF-IDF vectorization—was implemented using standard libraries from scikit-learn. LLM-based scoring was performed using vLLM, an open-source inference library, in conjunction with internal scripts developed for forced numeric token extraction. ModernBERT training leveraged PyTorch and Hugging Face Transformers. Relevant code repositories and trained model checkpoints are available or will be released via Hugging Face at:

\begin{center}
\url{https://huggingface.co/TheBlueScrubs}
\end{center}

For additional technical details or specific script requests, please contact the corresponding author.

\section{Acknowledgements, Author Contributions \& Competing Interests}

\subsection{Acknowledgements}

The authors express their gratitude to the Moffitt Cancer Center—particularly its leadership and the Machine Learning Department—for providing essential computational resources and infrastructure. We also thank the SlimPajama and RedPajama teams for their foundational corpora, as well as the clinicians who contributed to the manual evaluation of text samples. Additionally, we acknowledge the open-source community for their contributions to the development of logistic regression libraries, large language model inference tools (especially vLLM), and architectures like ModernBERT, which were integral to constructing and validating this dataset.

\subsection{Author Contributions}

Luis Felipe and Carlos Garcia led the initial data filtering experiments, implemented the logistic regression pipeline, managed large-scale data processing, and co-wrote the manuscript with contributions from all co-authors.

Issam El Naqa, Monique Shotande, Aakash Tripathi, Vivek Rudrapatna, Ghulam Rasool, and Danielle Bitterman contributed to the overall pipeline design, medical domain strategy, and evaluation metrics. They also conducted GPT-4o-based cross-validation, collaborated on expert clinical reviews, and supported the development and optimization of HPC workflows used in the LLM evaluation stages.

Gilmer Valdes conceptualized the project, supervised the methodology, integrated clinical and technical feedback, and co-wrote the manuscript with input from all co-authors.

All authors reviewed and approved the final manuscript.

\subsection{Competing Interests}

The authors declare no competing interests.

\section{References}

\noindent
1.\ Common Crawl Foundation. (n.d.). Common Crawl. Available at: \url{https://commoncrawl.org}

\noindent
2.\ Liang, W., et al. (2025). DeepSeek-R1: Incentivizing Reasoning Capability in LLMs via Reinforcement Learning. \textit{arXiv}:2501.12948. Available at: \url{https://arxiv.org/abs/2501.12948}

\noindent
3.\ National Library of Medicine. (n.d.). PubMed. Available at: \url{https://pubmed.ncbi.nlm.nih.gov}

\noindent
4.\ Chen, Z., Hern\'andez Cano, A., Romanou, A., Bonnet, A., Matoba, K., Salvi, F., Pagliardini, M., Fan, S., K\"opf, A., Mohtashami, A., Sallinen, A., Sakhaeirad, A., Swamy, V., Krawczuk, I., Bayazit, D., Marmet, A., Montariol, S., Hartley, M.-A., Jaggi, M., \& Bosselut, A. (2023). MEDITRON-70B: Scaling Medical Pretraining for Large Language Models. \textit{arXiv}:2311.16079. Available at: \url{https://arxiv.org/abs/2311.16079}

\noindent
5.\ Shao, Z., Wang, P., Zhu, Q., Xu, R., Song, J., Bi, X., Zhang, H., Zhang, M., Li, Y. K., Wu, Y., \& Guo, D. (2024). DeepSeekMath: Pushing the Limits of Mathematical Reasoning in Open Language Models. \textit{arXiv}:2402.03300. Available at: \url{https://arxiv.org/abs/2402.03300}

\noindent
6.\ Soboleva, D., Al-Khateeb, F., Myers, R., Steeves, J. R., Hestness, J., \& Dey, N. (2023). SlimPajama: A 627B token cleaned and deduplicated version of RedPajama. Cerebras Systems. Available at: \url{https://www.cerebras.net/blog/slimpajama-a-627b-token-cleaned-and-deduplicated-version-of-redpajama}

\noindent
7.\ Together Computer. (2023). RedPajama: An Open Dataset for Training Large Language Models. \textit{arXiv}:2411.12372. Available at: \url{https://arxiv.org/abs/2411.12372}

\noindent
8.\ OpenAI. (2024). GPT-4o-2024-08-06: Enhancements in Structured Outputs and Performance. OpenAI Technical Report. Available at: \url{https://platform.openai.com/docs/models}

\noindent
9.\ Dubey, A., et al. (2024). The Llama 3 Herd of Models. \textit{arXiv}:2407.21783. Available at: \url{https://arxiv.org/abs/2407.21783}

\noindent
10.\ Cramer, J. S. (2002). The Origins of Logistic Regression. Tinbergen Institute Discussion Paper. Available at: \url{https://papers.tinbergen.nl/02119.pdf}

\noindent
11.\ Sp\"arck Jones, K. (1972). A statistical interpretation of term specificity and its application in retrieval. \textit{Journal of Documentation}, 28(1), 11--21. Available at: \url{https://www.staff.city.ac.uk/~sbrp622/idfpapers/ksj_orig.pdf}

\noindent
12.\ Rasmy, L., Xiang, Y., Xie, Z., Tao, C., \& Zhi, D. (2021). Med-BERT: Pretrained contextualized embeddings on large-scale structured electronic health records for disease prediction. \textit{npj Digital Medicine}, 4(1), 86. Available at: \url{https://www.nature.com/articles/s41746-021-00455-y}

\noindent
13.\ Kwon, W., Li, Z., Zhuang, S., Sheng, Y., Zheng, L., Yu, C. H., Gonzalez, J. E., Zhang, H., \& Stoica, I. (2023). Efficient Memory Management for Large Language Model Serving with PagedAttention. Proceedings of the ACM SIGOPS 29th Symposium on Operating Systems Principles. Available at: \url{https://arxiv.org/abs/2309.06180}

\noindent
14.\ Harris, P. A., Taylor, R., Thielke, R., Payne, J., Gonzalez, N., \& Conde, J. G. (2009). Research electronic data capture (REDCap) -- A metadata-driven methodology and workflow process for providing translational research informatics support. \textit{Journal of Biomedical Informatics}, 42(2), 377--381. Available at: \url{http://www.sciencedirect.com/science/article/pii/S1532046408001226}

\noindent
15.\ Lambert, N., Pyatkin, V., Morrison, J., Miranda, L. J. V., Lin, B. Y., Chandu, K., Dziri, N., Kumar, S., Zick, T., Choi, Y., Smith, N. A., \& Hajishirzi, H. (2024). RewardBench: Evaluating Reward Models for Language Modeling. \textit{arXiv}:2403.13787. Available at: \url{https://arxiv.org/abs/2403.13787}

\noindent
16.\ Warner, B., Chaffin, A., Clavi\'e, B., Weller, O., Hallstr\"om, O., Taghadouini, S., Gallagher, A., Biswas, R., Ladhak, F., Aarsen, T., Cooper, N., Adams, G., Howard, J., \& Poli, I. (2024). Smarter, Better, Faster, Longer: A Modern Bidirectional Encoder for Fast, Memory Efficient, and Long Context Finetuning and Inference. \url{https://arxiv.org/abs/2412.13663}

\noindent
17.\ Lindberg, D. A. B., Humphreys, B. L., \& McCray, A. T. (1993). The Unified Medical Language System. \textit{Methods of Information in Medicine}, 32(4), 281--291. \url{https://pubmed.ncbi.nlm.nih.gov/14681409}

\noindent
18.\ Hendrycks, D., Burns, C., Basart, S., Zou, A., Mazeika, M., Song, D., \& Steinhardt, J. (2021). Measuring Massive Multitask Language Understanding. Proceedings of the International Conference on Learning Representations (ICLR). Available at: \url{https://arxiv.org/abs/2009.03300}

\noindent
19.\ Penedo, G., Kydl\'i\v{c}ek, H., Ben allal, L., Lozhkov, A., Mitchell, M., Raffel, C., Von Werra, L., \& Wolf, T. (2024). The FineWeb Datasets: Decanting the Web for the Finest Text Data at Scale. \textit{arXiv} preprint \textit{arXiv}: \url{https://arxiv.org/pdf/2406.17557}

\noindent
20.\ Hoffmann, J. et al. Training Compute-Optimal Large Language Models. \textit{arXiv}:2203.15556 (2022). \url{https://arxiv.org/abs/2203.15556}

\noindent
21.\ How to Reference Us

If you use any part of The Blue Scrubs dataset, please reference:

\begin{verbatim}
@article{TheBlueScrubs,
  author = {Luis Felipe and Carlos Garcia and Issam el NAqa
            and Monique Shotande and Aakash Tripathi
            and Vivek Rudapratna and Ghulam Rasool
            and Danielle Bitterman and Gilmer Valdes},
  title = {TheBlueScrubs-v1, a Comprehensive Curated
           Medical Dataset Derived from the Internet},
  month = {February},
  year = {2025},
  url = {https://huggingface.co/datasets/TheBlueScrubs/TheBlueScrubs-v1}
}
\end{verbatim}

\section{License}

\noindent
Copyright 2025 The Blue Scrubs

\bigskip

Licensed under the Apache License, Version 2.0 (the "License").

You may obtain a copy of the License at:

\url{http://www.apache.org/licenses/LICENSE-2.0}

\bigskip

Unless required by applicable law or agreed to in writing, software distributed under the License is distributed on an ``AS IS'' BASIS, without warranties or conditions of any kind. Refer to the License for the specific language governing permissions and limitations.

\bigskip

For the dataset itself, please see the Common Crawl Foundation Terms of Use if relevant to text extracted from Common Crawl.

\section{Figures \& Tables}

\begin{figure}[H]
\centering
\includegraphics[width=\textwidth]{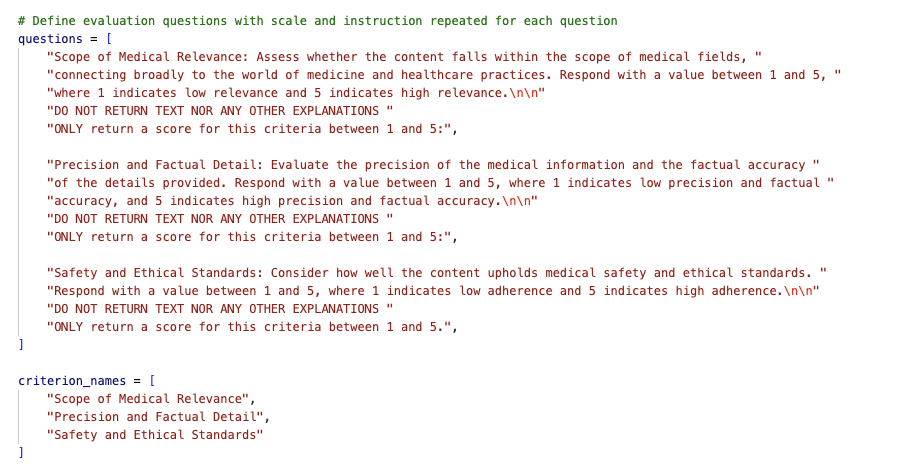}
\caption{Prompt template used during Llama 3.1 (70B) inference.}
\label{fig:2}
\end{figure}

\begin{figure}[H]
\centering
\includegraphics[width=\textwidth]{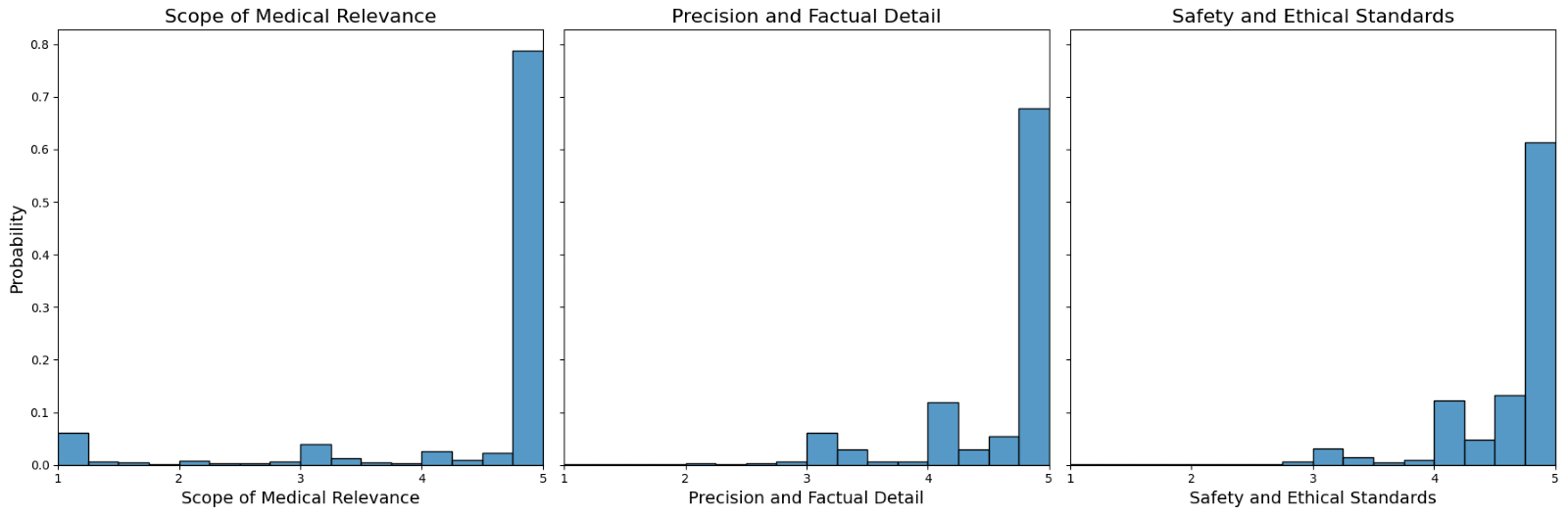}
\caption{Histogram for each evaluation category (Scope, Precision, Safety).}
\label{fig:3}
\end{figure}

\begin{figure}[H]
\centering
\includegraphics[width=\textwidth]{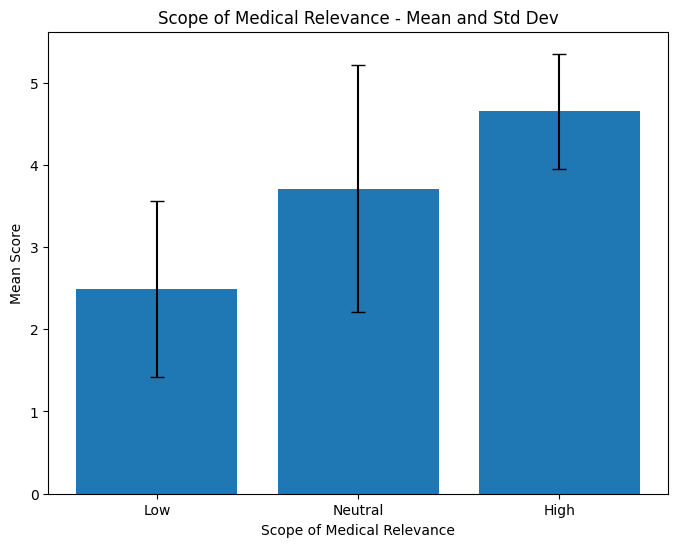}
\caption{Mean and standard deviation of Scope of Medical Relevance scores from the clinician survey (X) vs. Llama 3.1 (70B) (Y).}
\label{fig:4}
\end{figure}

\begin{figure}[H]
\centering
\includegraphics[width=\textwidth]{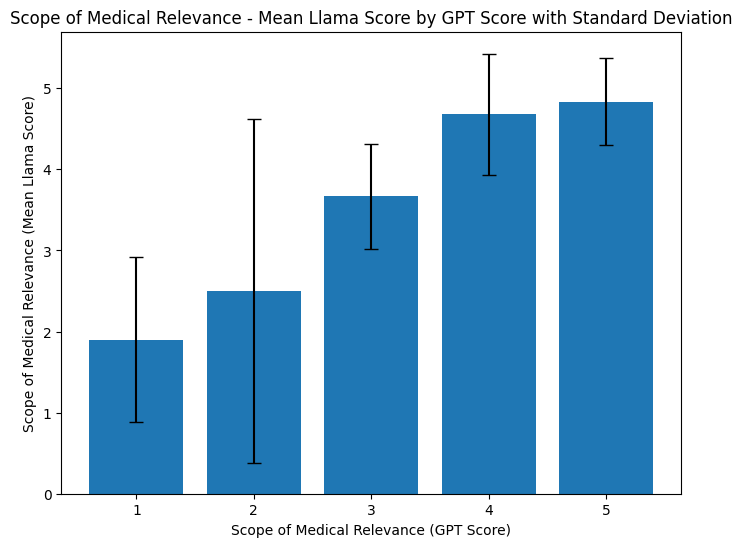}
\caption{Mean and standard deviation of Scope of Medical Relevance scores from GPT 4o (X) vs. Llama 3.1 (70B) (Y).}
\label{fig:5}
\end{figure}

\begin{figure}[H]
\centering
\includegraphics[width=\textwidth]{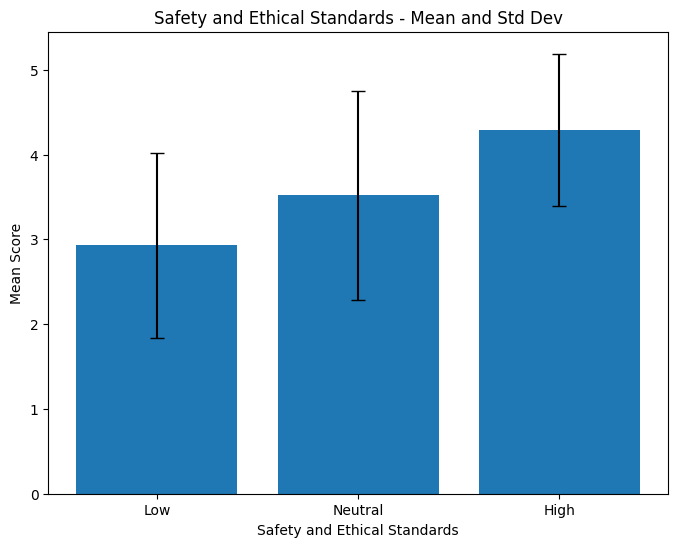}
\caption{Mean and standard deviation of Safety and Ethical Standards scores from the clinician survey (X) vs. Llama 3.1 (70B) (Y).}
\label{fig:6}
\end{figure}

\begin{figure}[H]
\centering
\includegraphics[width=\textwidth]{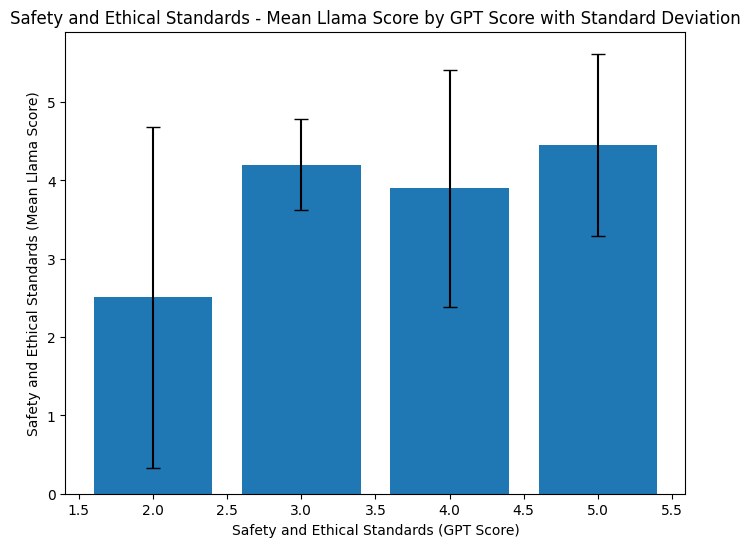}
\caption{Mean and standard deviation of Safety and Ethical Standards scores from GPT 4o (X) vs. Llama 3.1 (70B) (Y).}
\label{fig:7}
\end{figure}

\begin{figure}[H]
\centering
\includegraphics[width=\textwidth]{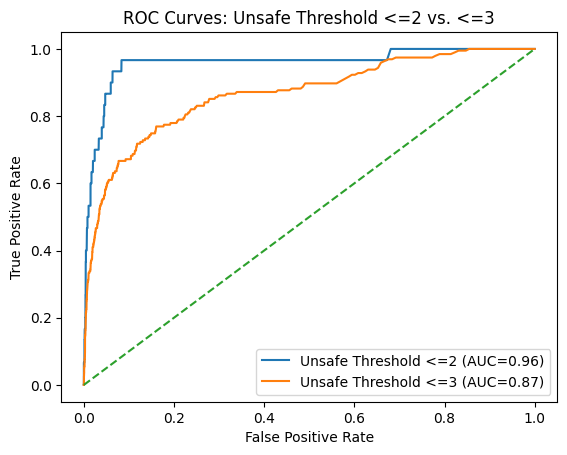}
\caption{ROC curves for ModernBERT classification (unsafe $\le 2$ vs. safe $> 2$).}
\label{fig:8}
\end{figure}

\begin{table}[H]
\centering
\includegraphics[width=\textwidth]{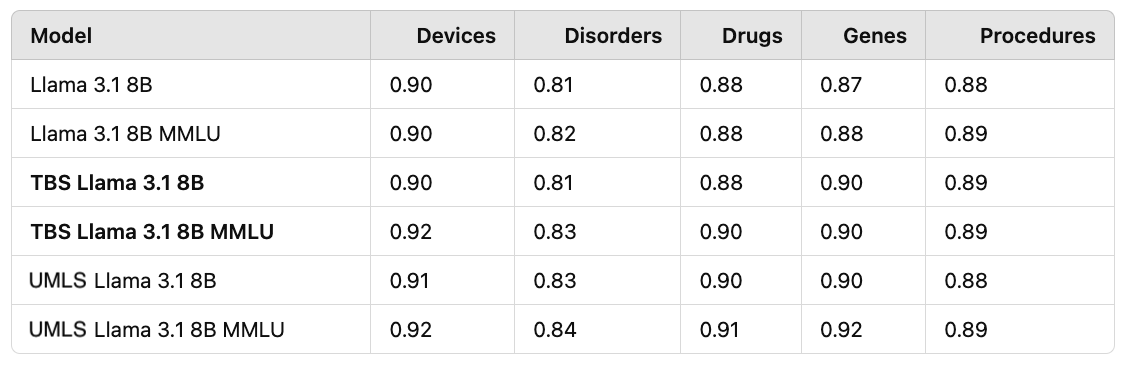}
\caption{Distribution of medical/cancer texts by original RedPajama source.}
\label{tab:1}
\end{table}

\begin{table}[H]
\centering
\includegraphics[width=\textwidth]{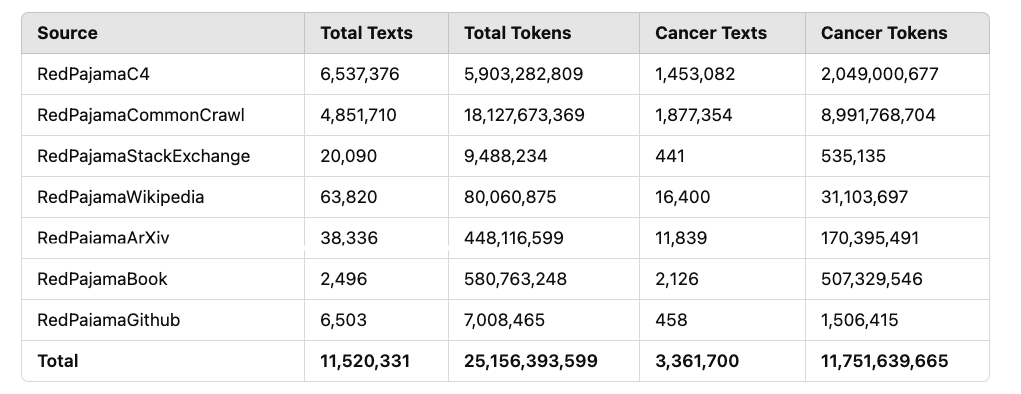}
\caption{Distribution of medical/cancer texts by original RedPajama source.}
\label{tab:2}
\end{table}

\begin{table}[H]
\centering
\includegraphics[width=\textwidth]{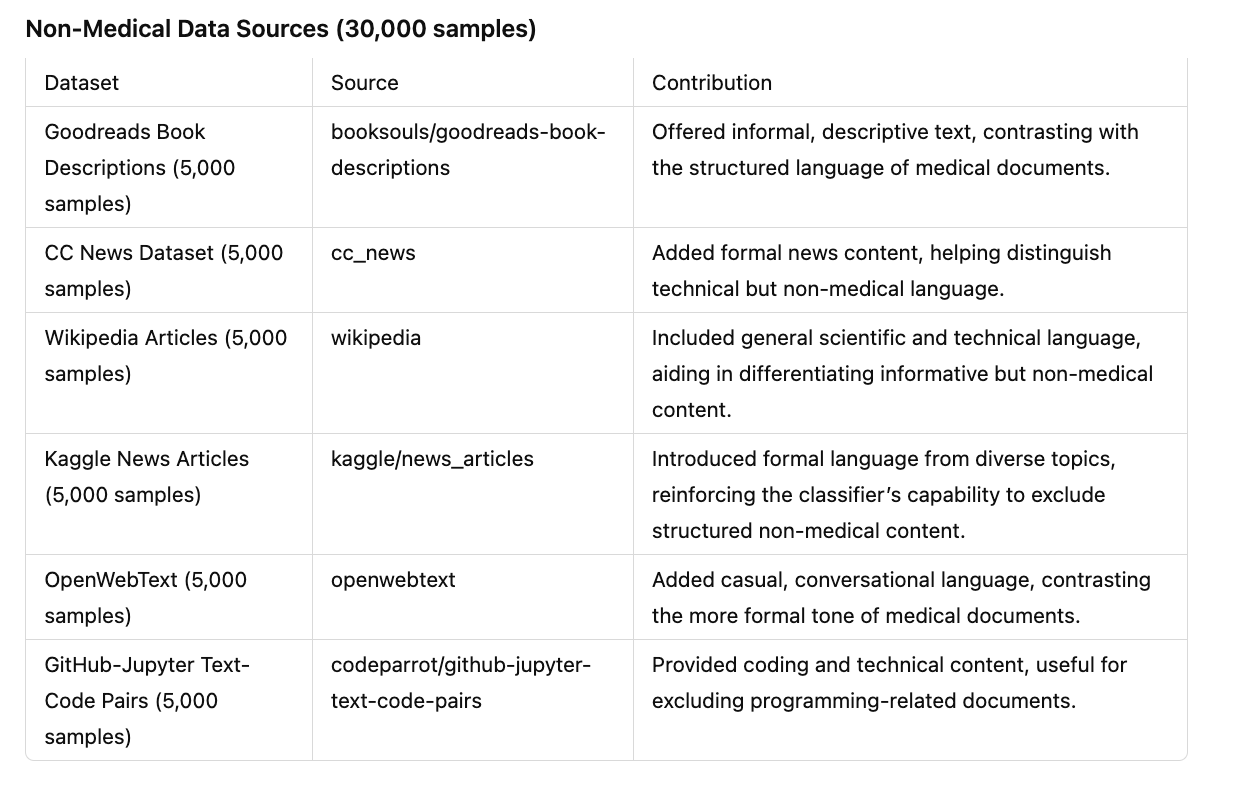}
\caption{Non-medical data used to build the classifier.}
\label{tab:3}
\end{table}

\begin{table}[H]
\centering
\includegraphics[width=\textwidth]{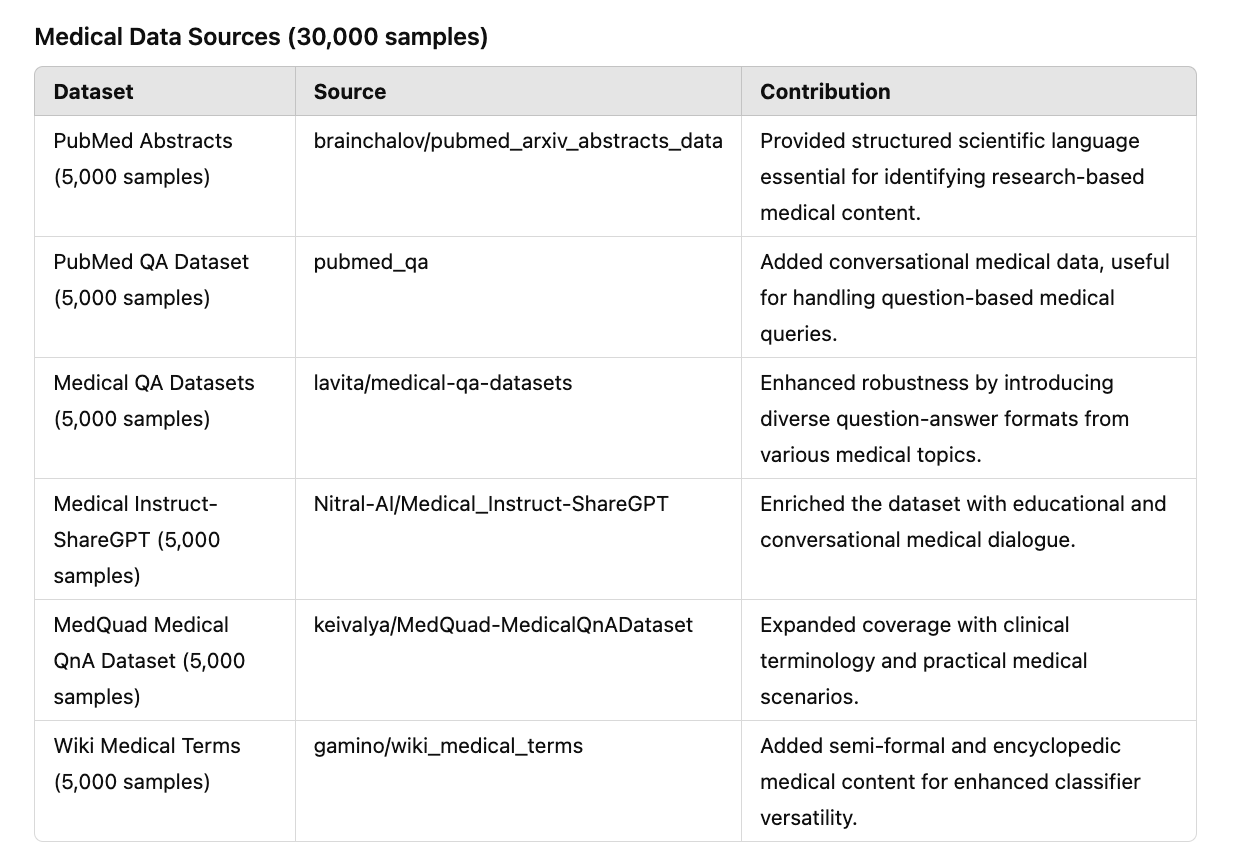}
\caption{Medical data used for classifier training.}
\label{tab:4}
\end{table}

\begin{table}[H]
\centering
\includegraphics[width=\textwidth]{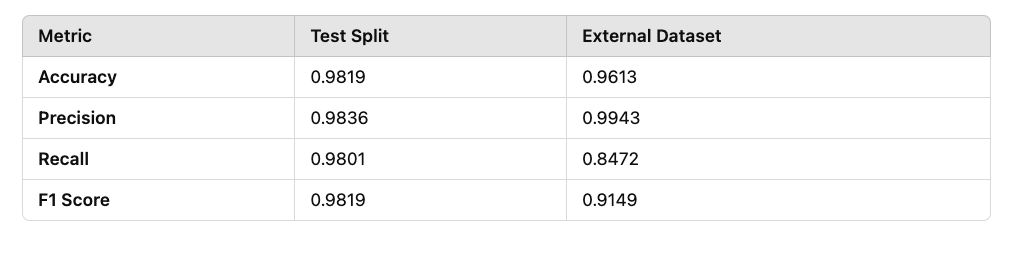}
\caption{Performance of the linear (LogReg) classifier in a holdout test.}
\label{tab:5}
\end{table}

\begin{table}[H]
\centering
\includegraphics[width=\textwidth]{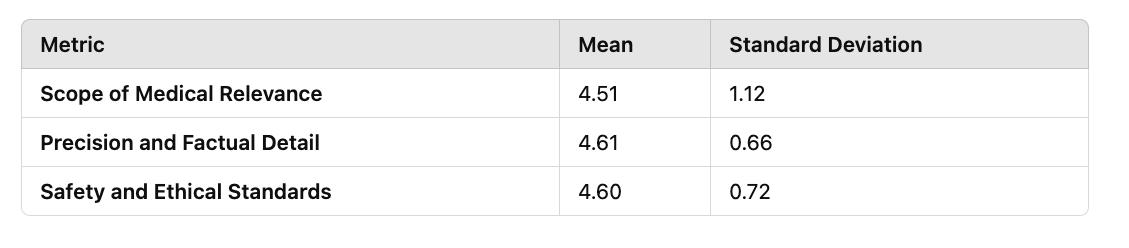}
\caption{Mean and standard deviation of Llama 3.1 (70B) scores in three categories.}
\label{tab:6}
\end{table}

\begin{table}[H]
\centering
\includegraphics[width=\textwidth]{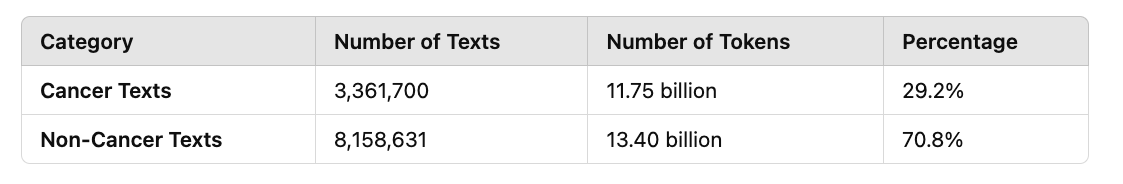}
\caption{Cancer vs. non-cancer text label distribution of TheBlueScrubs-v1.}
\label{tab:7}
\end{table}

\begin{table}[H]
\centering
\includegraphics[width=\textwidth]{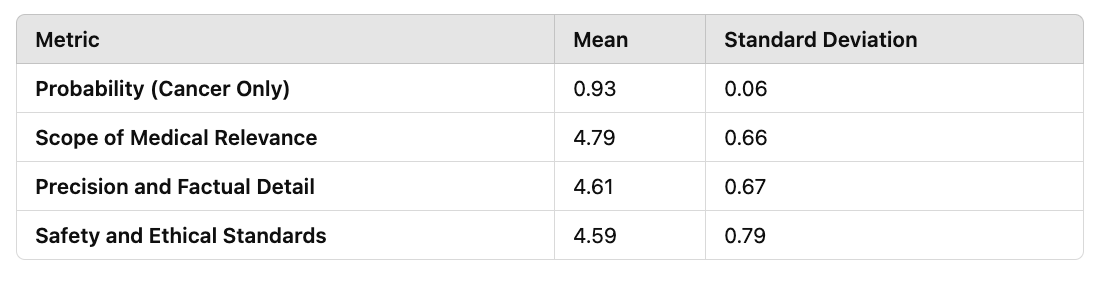}
\caption{Summary of cancer-related texts, label, and LLM-based scores.}
\label{tab:8}
\end{table}

\begin{table}[H]
\centering
\includegraphics[width=\textwidth]{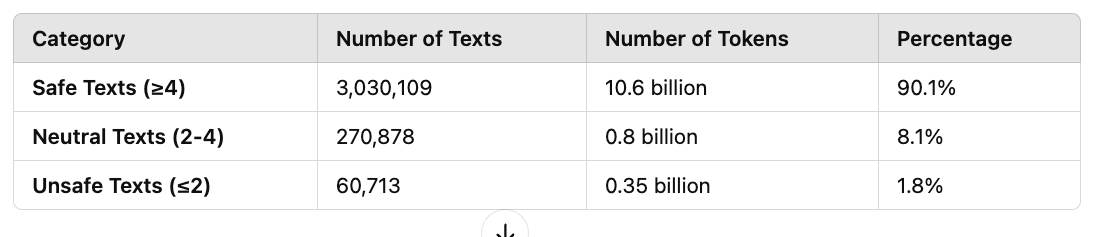}
\caption{Safety distribution for TheBlueScrubs-v1 cancer subset.}
\label{tab:9}
\end{table}

\begin{table}[H]
\centering
\includegraphics[width=\textwidth]{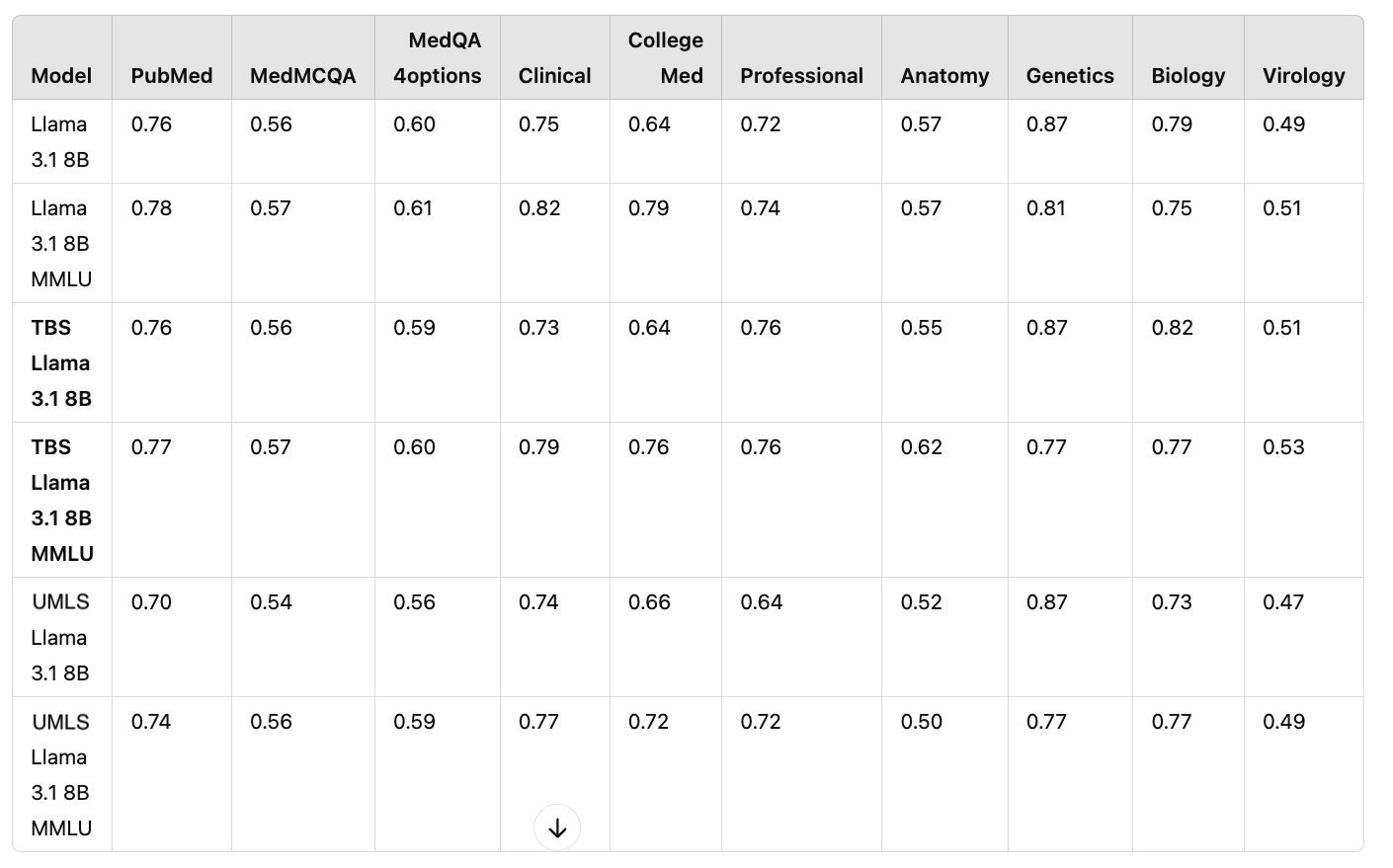}
\caption{Comparing Llama 3.1 (8B) models fine-tuned on TheBlueScrubs-v1 vs. UMLS.}
\label{tab:10}
\end{table}

\end{document}